\documentclass[10pt, conference, compsocconf]{IEEEtran}
\usepackage{cite}
\usepackage{hyperref}

\ifCLASSINFOpdf
\usepackage[pdftex]{graphicx}
\else
\fi
\usepackage[cmex10]{amsmath}
\usepackage{algorithmic}
\usepackage{url}
\usepackage[table,xcdraw]{xcolor}

\usepackage{fancyhdr}

\fancypagestyle{mystyle}{
  \fancyhf{} 
  \fancyhead[L]{Accepted to IEEE UIC 2024} 
  \fancyfoot[C]{\thepage}
}

\makeatletter
\def\ps@IEEEtitlepagestyle{
  \def\@oddfoot{\strut\hfill \raisebox{-5pt}{\parbox{\textwidth}{\centering
  \small © 2024 IEEE. Personal use of this material is permitted. Permission from IEEE must be obtained for all other uses, in any current or future media, including reprinting/republishing this material for advertising or promotional purposes, creating new collective works, for resale or redistribution to servers or lists, or reuse of any copyrighted component of this work in other works.}}}%
  \def\@evenfoot{}%
}
\makeatother

\begin{document}
%
\title{EDNet: Edge-Optimized Small Target Detection in UAV Imagery - Faster Context Attention, Better Feature Fusion, and Hardware Acceleration}




%

\DeclareRobustCommand*{\IEEEauthorrefmark}[1]{%
\raisebox{0pt}[0pt][0pt]{\textsuperscript{\footnotesize\ensuremath{#1}}}}
\author{
\IEEEauthorblockN{Zhifan Song\IEEEauthorrefmark{1,3}, 
Yuan Zhang\IEEEauthorrefmark{2}, 
Abd Al Rahman M. Abu Ebayyeh\IEEEauthorrefmark{3*}}
\IEEEauthorblockA{\IEEEauthorrefmark{1}LIP6 Laboratory, Sorbonne University, CNRS UMR7606, Paris, France}
\IEEEauthorblockA{\IEEEauthorrefmark{2}University of California, Berkeley, CA, United States}
\IEEEauthorblockA{\IEEEauthorrefmark{3}Department of Electrical and Electronic Engineering, Imperial College London, UK}
\IEEEauthorblockA{zhifan.song@lip6.fr, zhangyuan2024@berkeley.edu, a.abu-ebayyeh@imperial.ac.uk}
}


\maketitle

\begin{abstract}
Detecting small targets in drone imagery is challenging due to low resolution, complex backgrounds, and dynamic scenes. We propose EDNet, a novel edge-target detection framework built on an enhanced YOLOv10 architecture, optimized for real-time applications without post-processing. EDNet incorporates an XSmall detection head and a Cross Concat strategy to improve feature fusion and multi-scale context awareness for detecting tiny targets in diverse environments. Our unique C2f-FCA block employs Faster Context Attention to enhance feature extraction while reducing computational complexity. The WIoU loss function is employed for improved bounding box regression. With seven model sizes ranging from Tiny to XL, EDNet accommodates various deployment environments, enabling local real-time inference and ensuring data privacy. Notably, EDNet achieves up to a 5.6\% gain in mAP@50 with significantly fewer parameters. On an iPhone 12, EDNet variants operate at speeds ranging from 16 to 55 FPS, providing a scalable and efficient solution for edge-based object detection in challenging drone imagery. The source code and pre-trained models are available at: \url{https://github.com/zsniko/EDNet}.

\end{abstract}


%
\IEEEpeerreviewmaketitle

\section{Introduction}

The rapid advancement of commercial drones or unmanned aerial vehicles (UAVs) has brought transformative impacts across various sectors, including agriculture, aerial photography, shipping, security, and search and rescue~\cite{visdrone}. This growth has heightened the demand for accurate and efficient automated object detection. Moreover, the integration of UAVs equipped with advanced sensing technologies offers significant potential for real-time monitoring of social interactions and transportation dynamics, thereby enhancing social intelligence and urban management~\cite{bisio}.

\begin{figure}[htbp]
\centerline{\includegraphics[width=\linewidth]{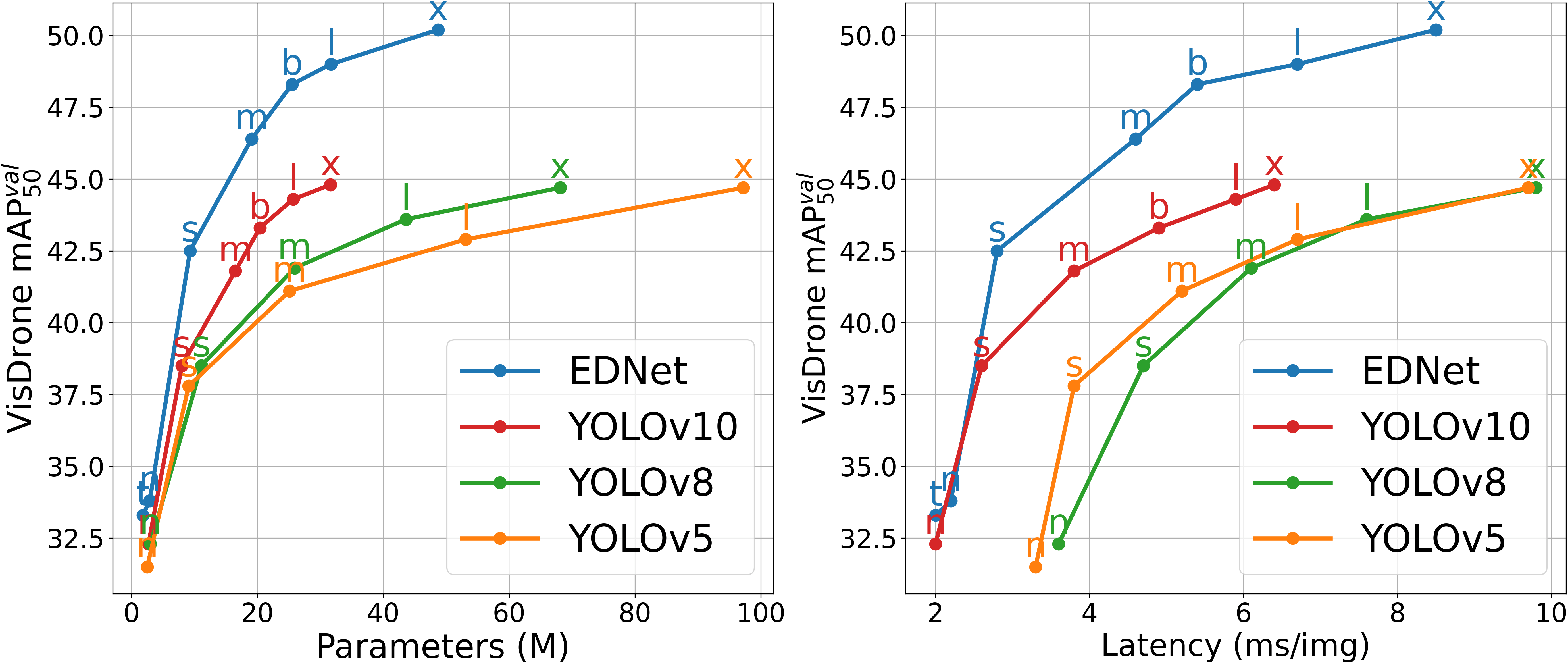}}
\caption{Comparison with state-of-the-art (SOTA) models for object detection. Size-mAP (left) and latency-mAP (right).}
\label{fig:benchmark}
\end{figure}

Recent advancements in deep learning, especially with the YOLO (You Only Look Once) \cite{yolo} family of one-shot object detectors, have significantly enhanced detection capabilities. Traditional two-stage detectors like Faster R-CNN \cite{fasterrcnn} excel in accuracy but fall short in speed. Transformers \cite{DETR} offer high accuracy but are more computationally intensive and less suitable for edge-device deployment. Previous YOLO versions relied on the Non-Maximum Suppression (NMS) post-processing technique for removing redundant bounding boxes. YOLOv10 \cite{yolov10}, the latest iteration of the YOLO series, eliminates post-processing by integrating dual label assignments and removing NMS, making it well-suited for real-time applications. Despite these advancements, detecting small objects in UAV imagery remains challenging due to resolution constraints and complex backgrounds, with objects of interest often comprise less than 10\% of the total pixel count~\cite{aodet}, compared to over 40\% in standard object detection datasets like MS COCO.

\begin{figure*}[htbp]
\centerline{\includegraphics[width=\textwidth]{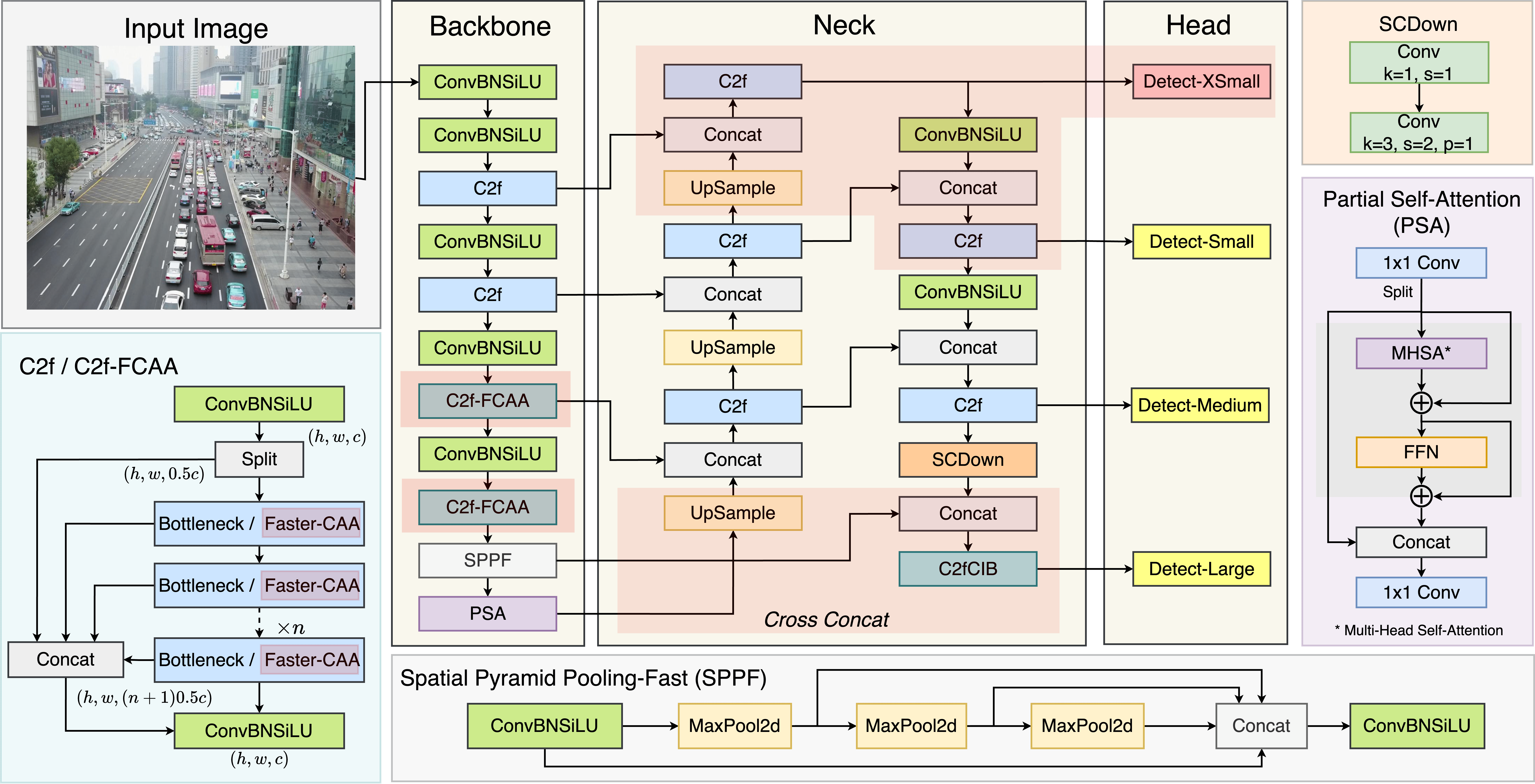}}
\caption{The proposed EDNet framework. The main architecture (backbone-neck-head) is illustrated in the center with a more detailed illustration of each block in the surroundings. ConvBNSiLU: Conv2d + Batch Normalization + SiLU.}
\label{fig:arch_ednet}
\end{figure*}

Recent advancements in UAV target detection using YOLO models have improved performance but present challenges for edge deployment. LV-YOLOv5~\cite{lvyolov5} integrates a vision transformer~\cite{vit} to enhance detection, but its higher parameter count limits edge applicability. Slicing-aided techniques with YOLOv5~\cite{slinet} and detection transformers~\cite{sahidetr} achieve high accuracy, but their heavy computational requirements make them unsuitable for edge computing. Various YOLOv5 improvements~\cite{v53}\cite{v54} have enhanced feature extraction, dynamic heads, data augmentation, and attention mechanisms, but did not address parameter efficiency. YOLOv8 models show gains through modifications to the C2f module and loss functions~\cite{v81}\cite{v82}, achieving a mean Average Precision (mAP) of 41.3\% but with increased complexity.

While some research has shifted toward lightweight models for UAV vision~\cite{light}, few have addressed the critical issue of edge deployment. VAMYOLOX~\cite{vamyolox}, a more efficient framework also available in seven sizes, achieves 47.7\% mAP@50 with its largest model but remains parameter-heavy and lacks investigation into edge-specific deployment. EdgeYOLO~\cite{edgeyolo}, designed specifically for parameter reduction, reaches up to 45.4\% mAP. However, its evaluation on embedded GPUs, with limited exploration of edge CPU performance, and the tiny variant still exceeds 5 million parameters.

Furthermore, related improved YOLO-based models intrinsically face speed limitations compared to the latest YOLOv10, which eliminates post-processing to streamline inference. To the best of our knowledge, there has been little exploration of YOLOv10 for UAV applications, presenting a gap in the literature. This makes YOLOv10 an ideal candidate as a baseline for optimizing a power-efficient and high-performance drone target detection framework tailored for edge deployment.

This paper introduces EDNet (EdgeDroneNet). Our primary contributions include:

\begin{itemize}
    \item Architectural Innovations: A novel C2f-FCA block featuring custom faster context attention for better feature extraction and reduced computational complexity; An additional XSmall detection head and Cross Concat Strategy (CCS) for better feature fusion; WIoUv3 for improved bounding box regression.

    \item Unmatched Performance: A framework spanning seven scalable variants (from Tiny to XL) consistently surpasses state-of-the-art object detectors with higher accuracy and unparalleled computational efficiency.

    \item Edge-Optimized Design: Hardware-accelerated models are tuned for seamless integration on mobile and edge devices. All variants operate in real-time on mobile devices like the iPhone 12, with an additional EDNet-Tiny variant specifically crafted for resource-constrained and low power platforms such as the Raspberry Pi.
    
\end{itemize}
\pagestyle{mystyle}
\section{Methodology}

The complete architecture of our proposed model (EDNet), is illustrated in Fig.~\ref{fig:arch_ednet}. The ConvBNSiLU (Conv2d, BatchNorm, SiLU) block, incorporating a 2D convolution, batch normalization, and Sigmoid Linear Unit (SiLU) activation~\cite{silu}, is a staple from YOLOv5~\cite{yolov5} and carries over to YOLOv8~\cite{yolov8} and YOLOv10~\cite{yolov10}. The Spatial-Channel Decoupled Downsampling (SCDown) block was first proposed in YOLOv10 and enhances efficiency by first using $1 \times 1$ convolution to adjust channel dimensions and then applying depthwise convolution for spatial downsampling, thus minimizing computational load while preserving crucial information. Attention mechanism is often used in object detection for better performance~\cite{revtransformer}, hence Partial Self-Attention (PSA)~\cite{yolov10} is proposed in the backbone, as a more computationally efficient alternative to traditional multi-head self-attention~\cite{Attention}, as depicted on the right side of Fig.~\ref{fig:arch_ednet}. Additionally, the Spatial Pyramid Pooling-Fast (SPPF)~\cite{sppf} layer, illustrated at the bottom of Fig.~\ref{fig:arch_ednet}, leverages three concatenated max-pooling operations to extract features at multiple scales. Subsequent sections will detail the specific enhancements made to the backbone, neck, and head.

\begin{figure*}[htbp]
\centerline{\includegraphics[width=\textwidth]{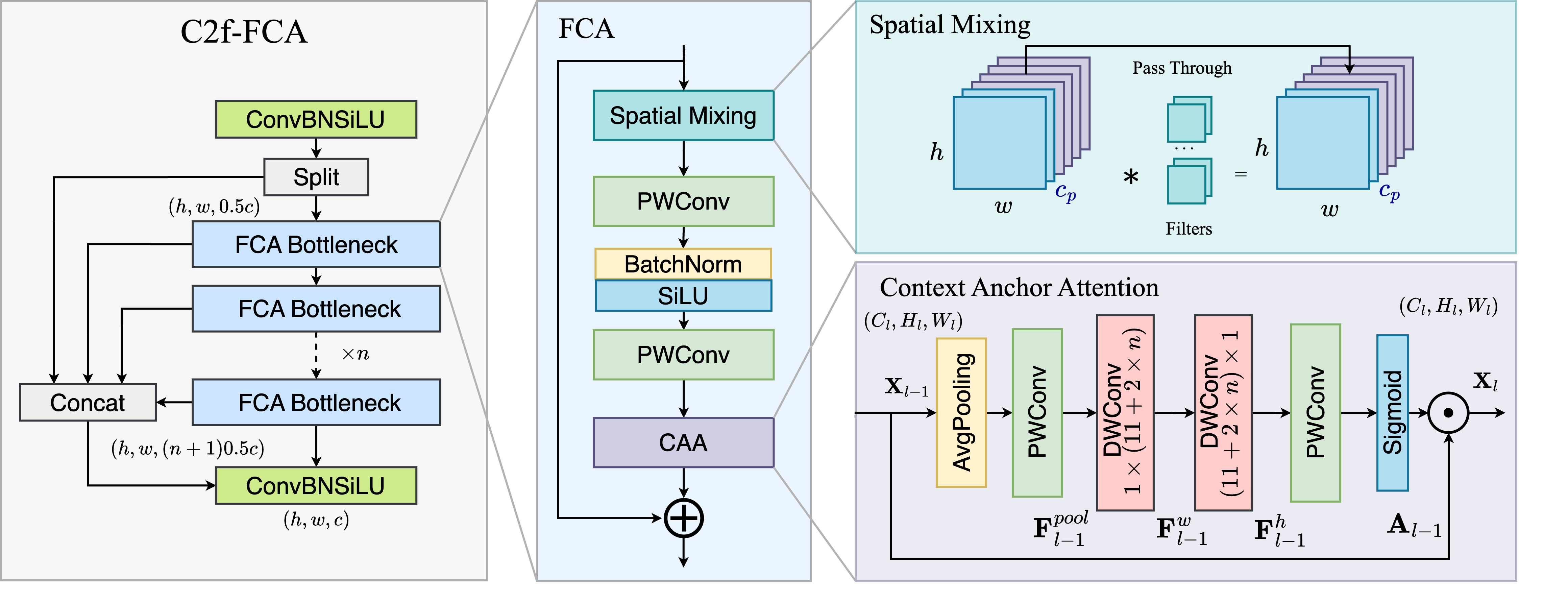}}
\caption{The proposed C2f-FCA block with Faster Context Attention bottleneck.}
\label{fig:c2ffcaa}
\end{figure*}

\subsection{Backbone Improvement}

We propose a novel and customized block, termed C2f-FCA, integrating an advanced Faster Context Attention (FCA) bottleneck. Our design starts with the FCA block, a sophisticated evolution of the FasterNet \cite{fasternet} architecture, which originally comprises a partial convolution followed by two $1 \times 1$ convolutions and a residual connection. This tailored configuration significantly enhances processing speed and efficiency. The Context Anchor Attention (CAA) is a sub-element derived from the Poly Kernel Inception Network \cite{pki}. Initially crafted for remote sensing applications, CAA is adeptly adapted here to improve feature extraction for drone-based target detection. SiLU \cite{silu} activation is also utilized to deliver smoother gradient transitions, promoting faster convergence and improving overall model performance. The FCA bottleneck not only improves performance but also reduces parameter count, making it more efficient. 

As illustrated in Fig. \ref{fig:c2ffcaa}, the FCA block incorporates $3 \times 3$ spatial mixing to blend spatial information from selected input channels, with drastically reduced computational complexity compared to traditional convolution. Additionally, it features a feed-forward network with two pointwise ($1 \times 1$) convolution (PWConv) layers and the CAA. This combination captures long-range contextual relationships among distant pixels, enhancing feature representation, especially in complex scenes with multiple objects of the same category. To achieve this, we first apply average pooling and a pointwise convolution to extract local features. Following this, we use two depth-wise strip convolutions to approximate the effect of large-kernel depth-wise convolutions in a computationally efficient manner due to its lightweight property and reduces the parameter by two for 2D traditional depth-wise convolution with size $k_{b}$. The attention mechanism employs a weighting matrix $\textbf{A}_{l-1}\in {\rm I\!R}^{C_{l} \times H_{l} \times W_{l}}$ to prioritize channel importance. The operations are summarized in Fig. \ref{fig:c2ffcaa} and are captured by the following equations:

\begin{equation}
    \textbf{F}_{l-1}^{pool} = \text{Conv}_{1\times 1}(\mathcal{P}_{avg}(\textbf{X}_{l-1})
\end{equation}

\begin{equation}
\begin{aligned}
    \textbf{F}_{l-1}^{w} &= \text{DWConv}_{1\times k_{b}}(\textbf{F}_{l-1}^{pool}), \\
    \textbf{F}_{l-1}^{h} &= \text{DWConv}_{k_{b}\times 1}(\textbf{F}_{l-1}^{w}).
\end{aligned}
\end{equation}

\begin{equation}
    \textbf{A}_{l-1} = \text{Sigmoid}(\text{Conv}_{1\times 1}(\textbf{F}_{l-1}^{h}))
\end{equation}

The CAA applies attention weighting to the channels based on their importance before Adding the original input back to the processed features, forming a skip connection that aids in training stability and performance. 

\subsection{Neck Improvement}
\subsubsection{XSmall Detection Head}
Drone cameras often capture vast scenes with tiny objects, presenting a challenge for effective target detection. The original YOLO downsamples feature maps through stages P1 to P5. For our input image size of 640, the resulting feature maps are 80 (P3), 40 (P4), and 20 (P5) pixels in resolution by the time they reach the detection heads. To improve the detection of small targets, we introduced an XSmall detection head, which features a resolution of $160 \times 160$ pixels. This head significantly reduces the down-sampling to just two stages, enabling it to retain more detailed and richer features of small targets. It is concatenated with features of the same scale from the backbone, as depicted in Fig. \ref{fig:arch_ednet}.  Adding the XSmall head to the neck of improves feature fusion by integrating finer-grained, high-resolution features into the detection process. In the subsequent section, we will demonstrate how this XSmall detection head enhances detection precision for tiny objects.

\subsubsection{Cross Concat - A Novel Feature Fusion Strategy}

We introduce a new concatenation scheme, Cross Concat Strategy (CCS), as depicted in Fig. \ref{fig:arch_ednet} to improve feature fusion in the detection process. Unlike YOLOv10, which forwards the PSA output to the first upsampling block and again to the last stage before the large detection head, our approach uses the SPPF output for cross-concatenation, while keeping the PSA output connected to the first upsampling block. The SPPF block pools feature maps at different scales, capturing rich multi-scale contextual information crucial for detecting objects of varying sizes in drone imagery. This adjustment may help mitigate the potential loss of broader context when using attention mechanisms late in the process, providing the final detection layers with a more comprehensive understanding of the scene. The performance gains observed, as discussed in the results section, suggest that Cross Concat could be a more effective strategy for target detection in complex aerial environments.

\subsection{Loss Function Improvement}
Drones often operate in diverse flying conditions, which can vary due to changes in altitude and interference noise. To address these challenges and enhance accuracy, we have opted to replace the conventional box loss function with the WIoUv3 loss \cite{wiou}. This choice enables more effective handling of noise by dynamically adjusting the focus across different samples, thus mitigating the impact of outliers.

\begin{equation}
L_{\text{WIoU}} = R_{\text{WIoU}} L_{\text{IoU}}
\end{equation}
\begin{equation}
R_{\text{WIoU}} = \exp\left(\frac{(x - x_{gt})^2 + (y - y_{gt})^2}{(W^2 + H^2)^*}\right)
\end{equation}

\begin{equation}
L_{\text{WIoUv3}} = r L_{\text{WIoU}}
\end{equation}
\begin{equation}
r = \frac{\beta}{\delta \alpha^{\beta - \delta}}; \quad
\beta = \frac{L_{\text{IoU}}^*}{\bar{L}_{\text{IoU}}} \in [0, +\infty)
\end{equation}

In these equations, ($x$, $y$) represents the coordinates of the ground truth bounding box, ($x_{gt}$, $y_{gt}$) are the predicted coordinate, and $W$ and $H$ denote the width and height of the minimal enclosing box between the two. The term 
$r$ is a non-monotonic focusing coefficient, 
$\beta$ quantifies the quality of outliers, and $\Bar{L_{\text{IoU}}}$is the moving average of the IoU loss.

By utilizing the WIoUv3 loss, we effectively reduce the influence of noisy data and outliers, leading to improved detection performance in challenging and variable drone environments.

\subsection{Hardware Acceleration}

In our focus on mobile computing, we developed an iOS application that deploys our model in CoreML format, facilitating seamless interaction between the CPU and the neural engine for accelerated AI inference.

Post-training quantization is applied with INT8 precision to minimize memory impact, while FP16 mixed precision is employed at runtime to maintain numerical accuracy. FP16 is also required by the neural engine for inference.

\begin{table}[htbp]
    \centering
    \normalsize
    \caption{System Configuration: Hardware and Software Environment}
    \begin{tabular}{ll}
        \hline
        \textbf{Category} & \textbf{Details} \\
        \hline
        GPU & NVIDIA A100 80GB PCIe \\
        CPU & Intel\textsuperscript{\textregistered} Xeon\textsuperscript{\textregistered} Gold 6300 @ 2 GHz \\
        RAM & 2 TB \\
        Operating System & Ubuntu 22.04 \\
        \hline
        Python Version  & 3.9.19 \\
        PyTorch Version & 2.0.1 \\
        CUDA Version & 11.8 \\
        \hline
    \end{tabular}
    \label{tab:setup}
\end{table}

Additionally, we optimized hardware performance by integrating industry-standard frameworks, including OpenVINO for Intel CPUs, TensorRT for NVIDIA GPUs, and ONNX for ARM-based edge processors for more generalized optimization. This approach optimizes the model's computational graph, ensuring efficient execution across diverse environments and enabling broad deployment scalability.

\section{Experimental Results}

\subsection{Dataset}
We utilized the VisDrone \cite{visdrone} dataset, a well-established and challenging benchmark for UAV-based object detection. The dataset comprises 6,471 training images and 548 validation images, encompassing 10 target categories: pedestrian, people, bicycle, car, van, tricycle, awning-tricycle, bus, and motorbike.

\begin{table*}[htbp]
    \centering
    \normalsize
    \caption{Performance comparison on VisDrone2019-DET-val against state-of-the-art models. YOLOv5, YOLOv8, and YOLOv10 do not have a tiny model variant. For larger models, only the top-performing YOLO versions are included. Latency was measured on A100 TensorRT FP32 to assess the upper bounds of performance in terms of accuracy and speed. Results from a few other models proposed in the literature are excluded from the table to ensure a fair comparison, as differences in hardware and software configurations can lead to variations in performance. These models are discussed in relevant sections.}
    \setlength{\tabcolsep}{12pt}
    \begin{tabular}{lcccc}
        \hline
        \textbf{Model} & \textbf{\#Param. (M) } & \textbf{mAP$_{50}$ (\%) } & \textbf{mAP$_{50-95}$ (\%) } & \textbf{Latency (ms)} \\
        \hline
        YOLOv3-Tiny~\cite{yolov3t} & 12.1 & 23.7 & 13.2 & 2.4 \\
        \rowcolor{gray!20} \textbf{EDNet-Tiny (Ours)} & \textbf{1.8} & \textbf{33.3} & \textbf{19.5} & 2.0 \\
        \hline
        YOLOv5-N~\cite{yolov5} & 2.5 & 31.5 & 18.1 & 3.3 \\
        YOLOv6-N~\cite{yolov6} & 4.2 & 29.8 & 17.4 & 3.8 \\
        YOLOv8-N~\cite{yolov8} & 3.0 & 32.3 & 18.7 & 3.6 \\
        YOLOv10-N~\cite{yolov10} & \textbf{2.7} & 32.3 & 18.8 & 2.0 \\
         \rowcolor{gray!20} \textbf{EDNet-N (Ours)} & 2.9 & \textbf{33.8} & \textbf{19.8} & 2.2 \\
        \hline
        YOLOv5-S~\cite{yolov5} & 9.1 & 37.8 & 22.4 & 3.8 \\
        YOLOv8-S~\cite{yolov8} & 11.1 & 38.5 & 22.9 & 4.7 \\
        YOLOv9-S~\cite{yolov9} & \textbf{7.2} & 39.0 & 23.4 & 4.7 \\ 
        YOLOv10-S~\cite{yolov10} & 8.0 & 38.5 & 22.9 & 2.6 \\
         \rowcolor{gray!20} \textbf{EDNet-S (Ours)} & 9.3 & \textbf{42.5} & \textbf{25.6} & 2.8 \\
        \hline
        YOLOv5-M~\cite{yolov5} & 25.1 & 41.1 & 25.0 & 5.2 \\
        YOLOv8-M~\cite{yolov8} & 25.9 & 41.9 & 25.4 & 6.1 \\
        YOLOv9-M~\cite{yolov9} & 20.0 & 43.1 & 26.1 & 7.3 \\ 
        YOLOv10-M~\cite{yolov10} & \textbf{16.5} & 41.5 & 25.4 & 3.8 \\
         \rowcolor{gray!20} \textbf{EDNet-M (Ours)} & 19.1 & \textbf{47.1} & \textbf{28.7} & 4.6 \\
        \hline
        YOLOv9-C~\cite{yolov9} & 25.3 & 43.1 & 26.4 & 6.3 \\
        YOLOv10-B~\cite{yolov10} & \textbf{20.4} & 43.4 & 26.6 & 4.9 \\
        \rowcolor{gray!20} \textbf{EDNet-B (Ours)} & 25.5 & \textbf{48.3} & \textbf{29.9} & 5.4 \\
        \hline
        YOLOv5-L~\cite{yolov5} & 53.1 & 42.9 & 26.2 & 6.3 \\ 
        YOLOv8-L~\cite{yolov8} & 43.6 & 43.6 & 26.8 & 7.6 \\
        YOLOv10-L~\cite{yolov10} & \textbf{25.7} & 44.3 & 27.1 & 5.9 \\
        RT-DETR-L~\cite{rtdetr} & 32.0 & 38.1 & 21.8 & 6.7 \\ 
         \rowcolor{gray!20} \textbf{EDNet-L (Ours)} & 31.7 & \textbf{49.0} & \textbf{30.4} & 6.7 \\
        \hline
        YOLOv5-X~\cite{yolov5} & 97.2 & 44.7 & 27.4 & 9.7 \\ 
        YOLOv8-X~\cite{yolov8} & 68.1 & 44.7 & 27.6 & 9.8 \\
        YOLOv9-E~\cite{yolov9} & 57.4 & 46.5 & 28.9 & 11.8 \\ 
        YOLOv10-X~\cite{yolov10} & \textbf{31.6} & 44.8 & 27.6 & 6.4 \\
        RT-DETR-X~\cite{rtdetr} & 65.5 & 40.8 & 23.6 & 8.9 \\
         \rowcolor{gray!20} \textbf{EDNet-X (Ours)} & 48.7 & \textbf{50.2} & \textbf{31.2} & 8.5 \\
        \hline
    \end{tabular}
    \label{tab:model_comparison}
\end{table*}

\subsection{Model Training}
\subsubsection{Environment Setup}
The details of the hardware and software configurations are outlined in Table \ref{tab:setup}. Stochastic Gradient Descent (SGD) is employed as the optimizer, with a learning rate set to 0.01 and a momentum of 0.9. All models are trained for 200 epochs to ensure full convergence, and the best-performing model is selected during the training process.

\subsubsection{Evaluation Metrics}

In object detection tasks, we utilize standard performance metrics: precision, recall, and mean Average Precision (mAP).

\begin{equation}
\text{Precision (P)} = \frac{\text{TP}}{\text{TP + FP}}
\end{equation}

\begin{equation}
\text{Recall (R)} = \frac{\text{TP}}{\text{TP + FN}}
\end{equation}

\begin{equation}
\text{Average \ Precision (AP)}= \int_{0}^{1} p(r) , dr
\end{equation}

\begin{equation}
\text{mean \ Average \ Precision (mAP)}= \frac{1}{k} \sum_{i=1}^{k}AP_{i}
\end{equation}

Where p(r) is the precision as a function of recall. The mAP is computed by averaging the AP values across multiple classes. In these equations, TP represents true positives, FP stands for false positives, TN denotes true negatives, and FN indicates false negatives.

\subsection{Model Comparison and Discussion}

\begin{figure*}[htbp]
\centerline{\includegraphics[width=\textwidth]{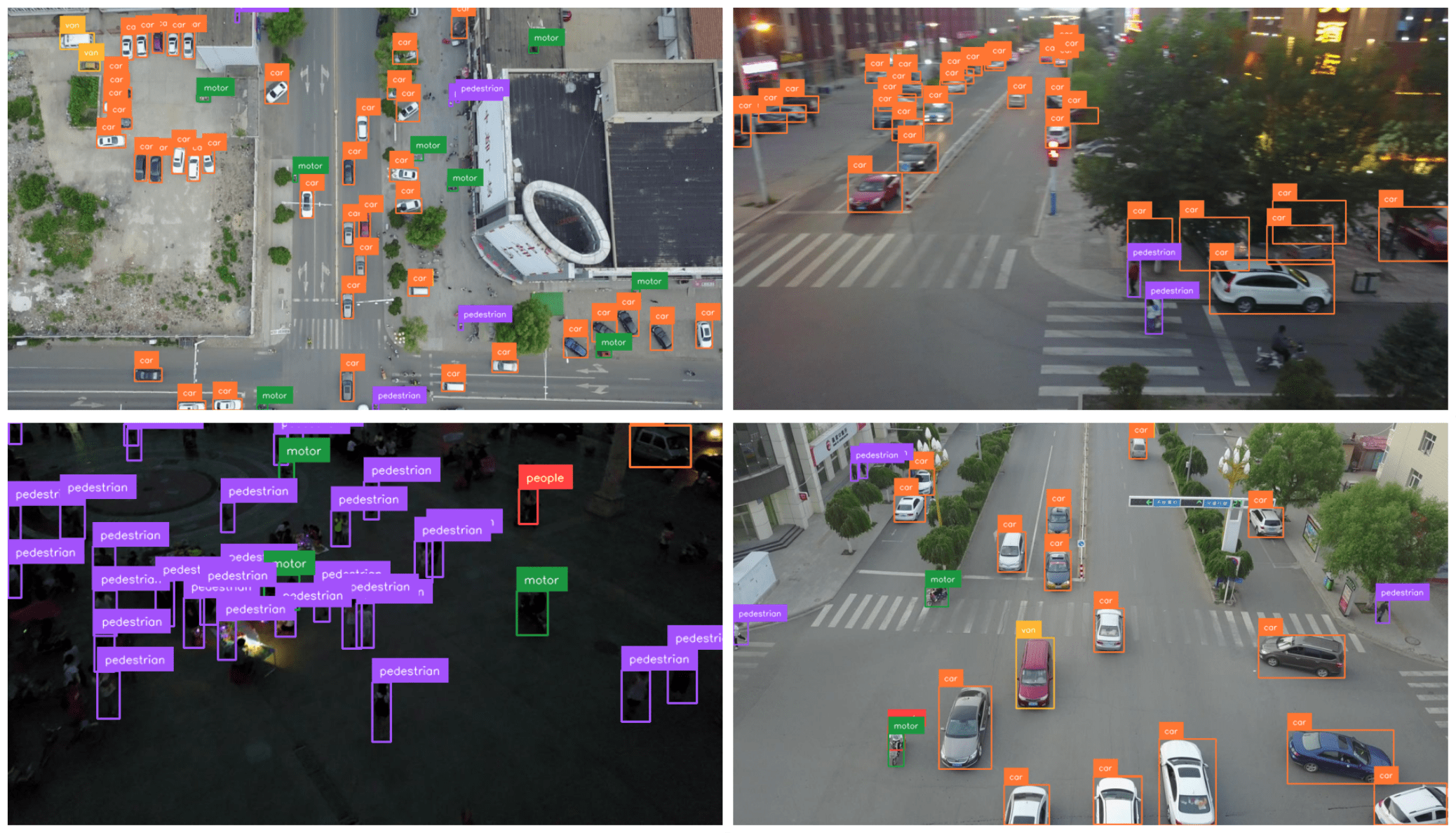}}
\caption{Sample predictions with the 1.78M EDNet-Tiny model under various scenarios.}
\label{fig:pred}
\end{figure*}

We evaluated the performance of EDNet against SOTA models, especially with variants also existing in YOLOv10: N, S, M, B, L, and X. To further optimize for edge deployment, we introduce a new tiny variant by reducing the depth scale factor from 0.33 to 0.2 and capping the maximum number of channels from 1024 to 512. This results in a model with just 1.78M parameters, significantly fewer than any YOLO variant, making it uniquely suited for deployment on resource-constrained devices, such as Raspberry Pi, without notably compromising detection performance. While small and medium variants could operate efficiently on modestly powered edge CPUs, larger variants may require more robust embedded GPUs, such as NVIDIA Jetson devices. This adaptability across a spectrum of hardware configurations underscores EDNet's scalability and practicality for diverse deployment scenarios, meeting both performance and operational requirements in real-world applications.

As demonstrated in Table \ref{tab:model_comparison}, EDNet consistently outperforms its competitors across all model sizes. Despite YOLOv3-Tiny having more parameters than EDNet-Tiny, and even exceeding EDNet-S and EDNet-N, it delivers inferior mAP. Furthermore, models like YOLOv6-N underperform compared to YOLOv5-N, YOLOv8-N, and YOLOv10-N, and were thus excluded from further consideration. Although YOLOv9 performs well, its slower inference speeds make it less suited for edge applications, solidifying YOLOv5, YOLOv8, and YOLOv10 as the primary YOLO-based competitors in our analysis.

To provide a broader perspective, we also compared EDNet to Real-Time Detection Transformer (RT-DETR)~\cite{rtdetr}, the only transformer-based model capable of real-time target detection, which is critical for edge computing. However, while RT-DETR exhibits strong performance in general object detection, it struggles with small object detection, a key requirement in drone imagery. For example, RT-DETR-X achieves mAP@50 of 40.8\%, which is 6.3\% lower than EDNet-M, despite using 70.8\% more parameters, making it less ideal for drone imagery on edge computing.

\begin{figure*}[htbp]
\centerline{\includegraphics[width=\textwidth]{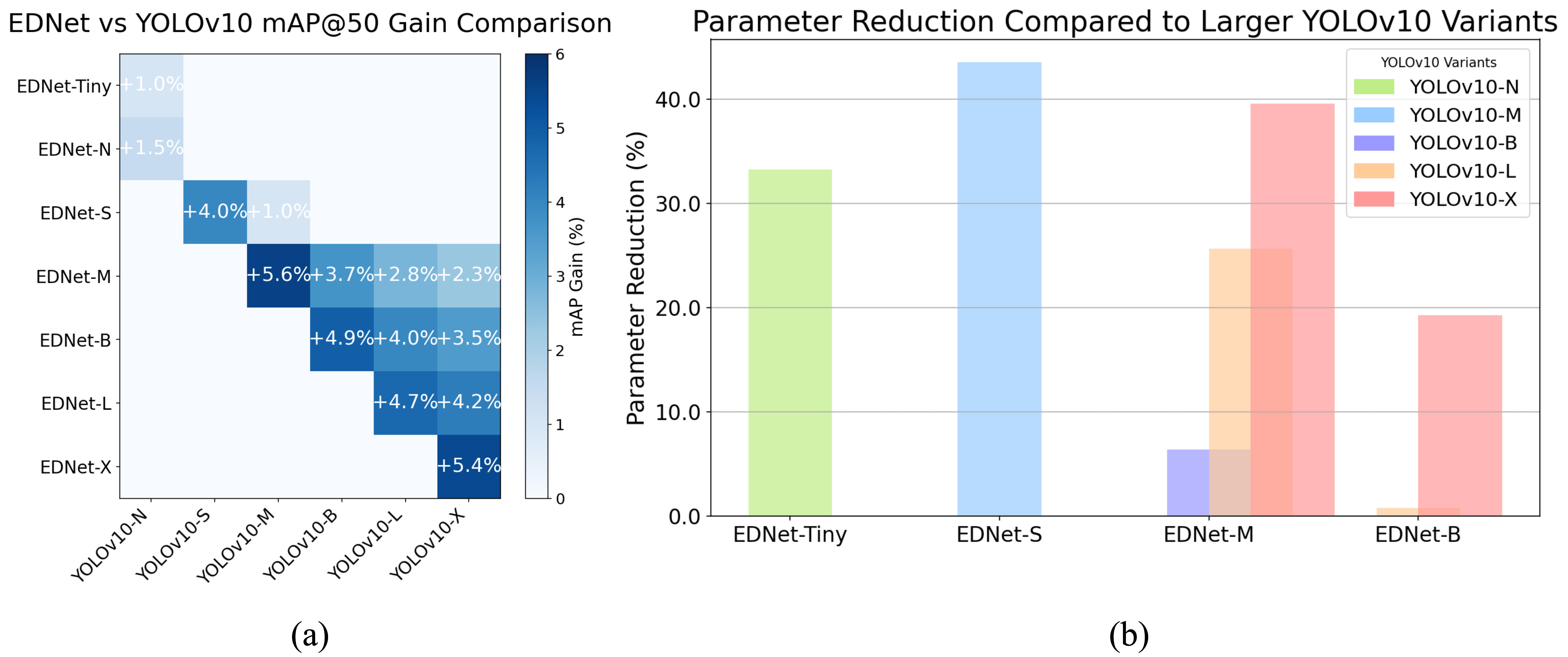}}
\caption{Performance comparison between EDNet and YOLOv10: (a) mAP gain relative to models of equal or larger size; (b) Parameter reduction compared to larger YOLOv10 models while achieving higher mAP.}
\label{fig:comparemod}
\end{figure*}

The comparison results presented in Fig.~\ref{fig:benchmark} unequivocally demonstrate EDNet's substantial performance superiority over YOLOv10 across various configurations. As depicted in Fig.~\ref{fig:comparemod} (a), EDNet consistently surpasses YOLOv10 in mAP@50, particularly evident when comparing equivalent model sizes along the diagonal and in larger YOLOv10 variants. Notably, EDNet begins to outperform YOLOv10 at larger variants, specifically starting from the S variant, and continues to exceed all YOLOv10 configurations from EDNet-M onward, including YOLOv10-X. Additionally, Fig.~\ref{fig:benchmark} and Fig.~\ref{fig:comparemod} (b) also illustrates that EDNet achieves comparable performance while utilizing significantly fewer parameters. For example, EDNet-S is 43.6\% smaller than YOLOv10-M while delivering 1\% better performance. Furthermore, EDNet-M outperforms the largest YOLOv10-X by 2.3\% in mAP@50 while requiring 39.6\% fewer parameters. The difference in parameter count is even more striking when compared to YOLOv8-X, with EDNet utilizing 71.9\% fewer parameters. Moreover, EDNet offers a remarkably compact version, EDNet-Tiny, which incorporates just 1.78 million parameters and achieves an impressive mAP@50 of 33.3\%, surpassing all N variants of SOTA YOLO models. These findings not only affirm EDNet’s superior detection accuracy but also underscore its efficiency, as it achieves similar mAP scores with a fraction of the parameters. 

In a comparative analysis of some efficient models optimized for drone imagery, VAMYOLOX-X~\cite{vamyolox} achieves a mAP@50 of 47.6\% with 104.6 million parameters. This represents a substantial 53\% increase in parameter count compared to EDNet-X, which, despite having fewer parameters, delivers higher mAP@50 of 50.2\%. Furthermore, when evaluating other edge-optimized object detection models, such as EdgeYOLO~\cite{edgeyolo}, we note that the EdgeYOLO-Tiny variant contains 5.5 million parameters and is benchmarked on NVIDIA Jetson, making it relatively resource-intensive for CPU-based edge devices. In contrast, EDNet-M not only demonstrates superior performance—achieving a 2.3\% improvement in mAP@50 over the largest EdgeYOLO variant while utilizing 53\% fewer parameters. 

These findings position EDNet as a superior model for real-time applications and edge deployments. Its precision, scalability, and efficiency make it a powerful tool for UAV imagery and small target detection, outperforming both YOLO and transformer-based models in key performance metrics while offering unmatched resource efficiency. Performance test results from real-world deployment scenarios will be discussed in Section \ref{sec:deploy}.

Fig. \ref{fig:pred} presents sample predictions using the EDNet-Tiny variant. This figure showcases various challenging scenarios encountered by drones, including diverse viewing angles, blurred scenes (top right), and low-light conditions (bottom left). Despite the substantial reduction in parameters for edge device optimization, the EDNet-Tiny variant demonstrates impressive performance in accurately detecting targets across these diverse and demanding environments. This robustness underscores the effectiveness of the EDNet-Tiny in maintaining high detection quality even under constrained conditions, highlighting its suitability for real-world edge deployments.

\subsection{Ablation Experiment}

We conducted the ablation study, each incremental addition demonstrates significant performance enhancements, affirming the effectiveness of our architectural choices. We pick EDNet-b to illustrate the ablation results in Table \ref{tab:ablation}, due to previously claimed reasons. 
\begin{figure}[htbp]
\centerline{\includegraphics[width=\linewidth]{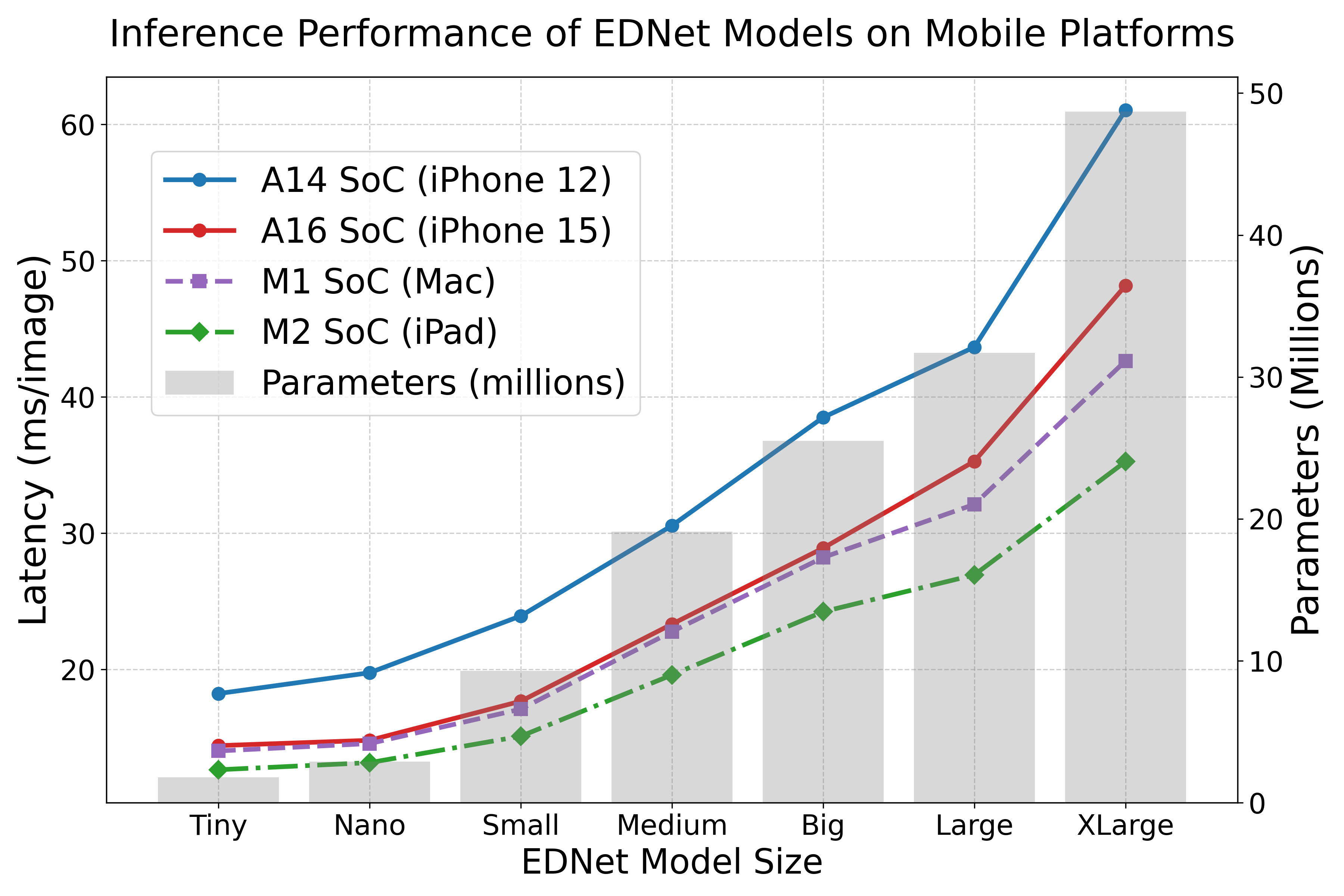}}
\caption{Inference performance of seven EDNet variants on mobile platforms. Lower latency is better.}
\label{fig:mobile}
\end{figure}
The introduction of the WIoU loss led to a notable improvement in bounding box regression, effectively handling noisy and low-quality samples. This refinement alone contributed to a 0.3\% increase in mAP@50. The addition of the XSmall detection head marked a substantial leap forward, especially in the detection of tiny targets. Further integrating the CCS into the model significantly improved feature fusion within the neck architecture. The neck enhancement resulted in a remarkable 4.2\% increase in mAP@50 and a 3.0\% increase in mAP@50-95, elevating the model's performance to 47.9\% and 29.7\%, respectively. This gain underscores the critical role of multi-scale detection and the importance of enhancing feature connectivity and information flow for better detection precision. Nevertheless, the extra detection head lead to an increase of computational overhead. 

\begin{table*}[htbp]
    \normalsize
    \centering
    \caption{Ablation experiment}
    \setlength{\tabcolsep}{18pt}
    \begin{tabular}{lccc}
        \hline
        \textbf{Model} &\textbf{\#Param} & \textbf{mAP$_{50}$ } & \textbf{mAP$_{50:95}$} \\
        & (M) & (\%) & (\%) \\ 
        \hline
        Baseline & 20.4 & 43.4\% & 26.6\%\\
        EDNet-B (WIoU) & 20.4 & 43.7\% & 26.7\%   \\
        EDNet-B (WIoU + XSmallHead) & 29.9 & 47.4\% & 29.5\% \\
        EDNet-B (WIoU + XSmallHead + CCS) & 29.9 & 47.9\% & 29.7\% \\
        EDNet-B (WIoU + XSmallHead + CCS + C2f-FCA) & \textbf{25.5} & \textbf{48.3\%} & \textbf{29.9\%} \\
        \hline
    \end{tabular}
    \label{tab:ablation}
\end{table*}

The final incorporation of the C2f-FCA block proved to be a game-changer. Not only did it optimize feature extraction, resulting in the highest recorded performance of 48.3\% mAP@50 and 29.9\% mAP@50-95, but it also addressed the increased model complexity from the XSmall detection head. Remarkably, it reduced the model's parameters from 29.9M to 25.5M while improving speed and accuracy. The complete EDNet model represents an overall gain of 4.9\% in mAP@50 and 3.3\% in mAP@50-95 compared to the baseline.

\subsection{Model Deployment}
\label{sec:deploy}

\begin{figure}[t]
\centerline{\includegraphics[width=\linewidth]{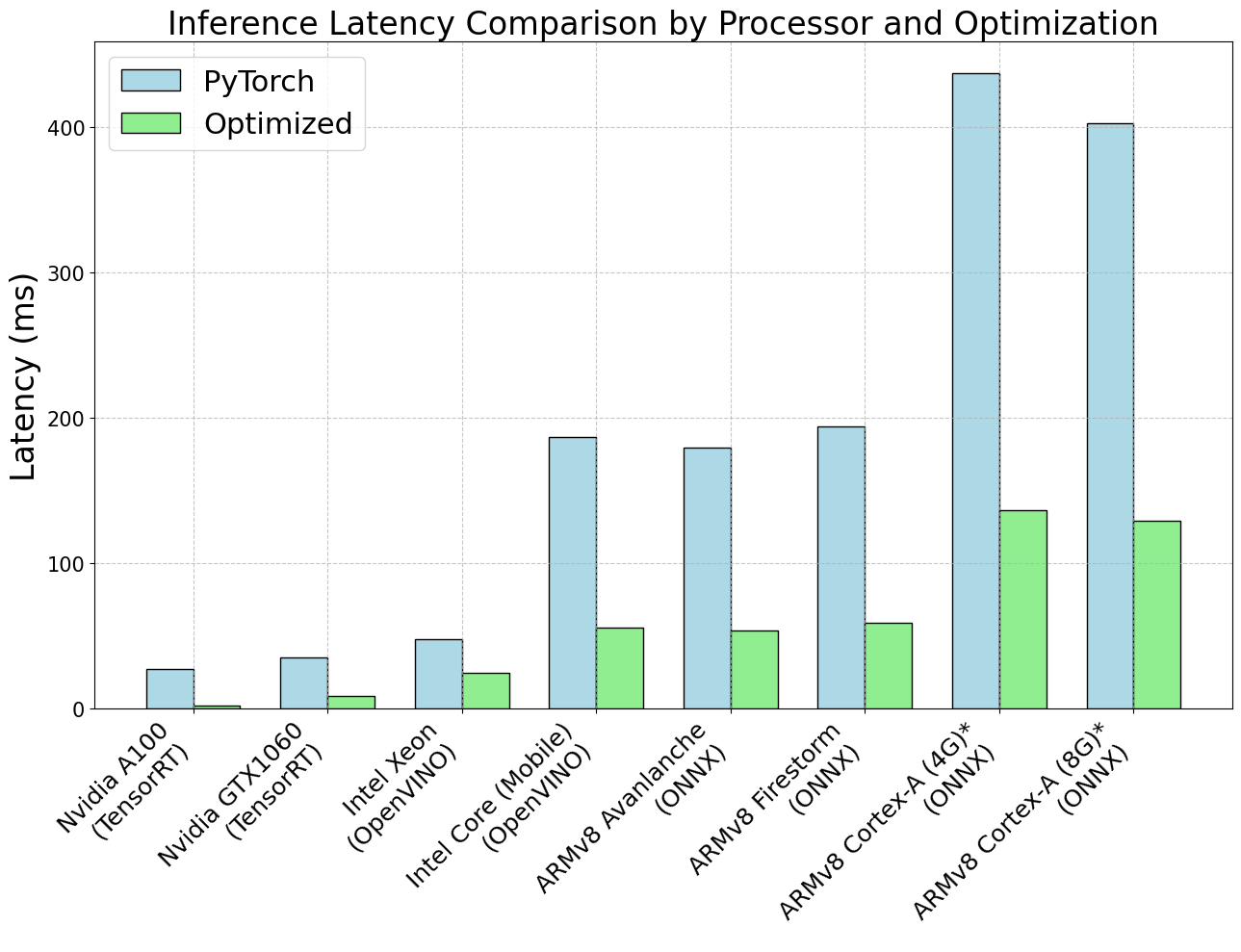}}
\caption{Comparison of inference speed in milliseconds before and after optimization (* Raspberry Pi 5).}
\label{fig:optim}
\end{figure}

We investigate the deployability of EDNet-Tiny across various environments within a generalized deployment setup, with an emphasis on ARM-based CPUs, which are prevalent in mobile devices. The model demonstrates a remarkable 3.2-fold increase in speed on edge processors, achieving latencies of 53.4 ms and 59.1 ms per image on the ARMv8 Avalanche and Firestorm performance cores, respectively, compared to its raw implementation with PyTorch on the same processors. Notably, on an ARMv8 Cortex-A76 processor within a Raspberry Pi 5, an exemplar of a resource-constrained embedded device, EDNet-Tiny processes images 3.1 times faster, with latencies of 129 ms versus 402 ms for the raw PyTorch model. Additionally, the 8GB version outperforms the 4GB version slightly, underscoring the importance of memory capacity in optimizing performance. Results illustrating latency improvements before and after optimization are presented in Fig.~\ref{fig:optim}, highlighting the speed enhancements achieved through this process.

We further evaluate the inference performance of EDNet in the context of specialized mobile computing, as mobile phones can be conveniently integrated into drones for image capture. Drones can also be fully operated using onboard smartphones, which serve dual functions for vision and control~\cite{phonedrone}. The pre-trained EDNet models are implemented via the CoreML framework, facilitating accelerated AI inference by interfacing the CPU and the neural engine. All seven model sizes, ranging from Tiny to XL, are tested on the A14 System On Chip (SoC) of an iPhone 12 (released in 2020). Additionally, we assess the performance on the A16 SoC (2022) in an iPhone 15, the M1 SoC (2020) in a Mac and the M2 SoC (2022) in an iPad, which possess greater computational power that could approximate the capabilities of more recently released mobile devices. Notably, the performance curve for the iPhone 15 (red) is already approaching that of the M1 Mac (blue), as shown in Fig.~\ref{fig:mobile}.

\begin{figure*}[htbp]
\centerline{\includegraphics[width=\textwidth]{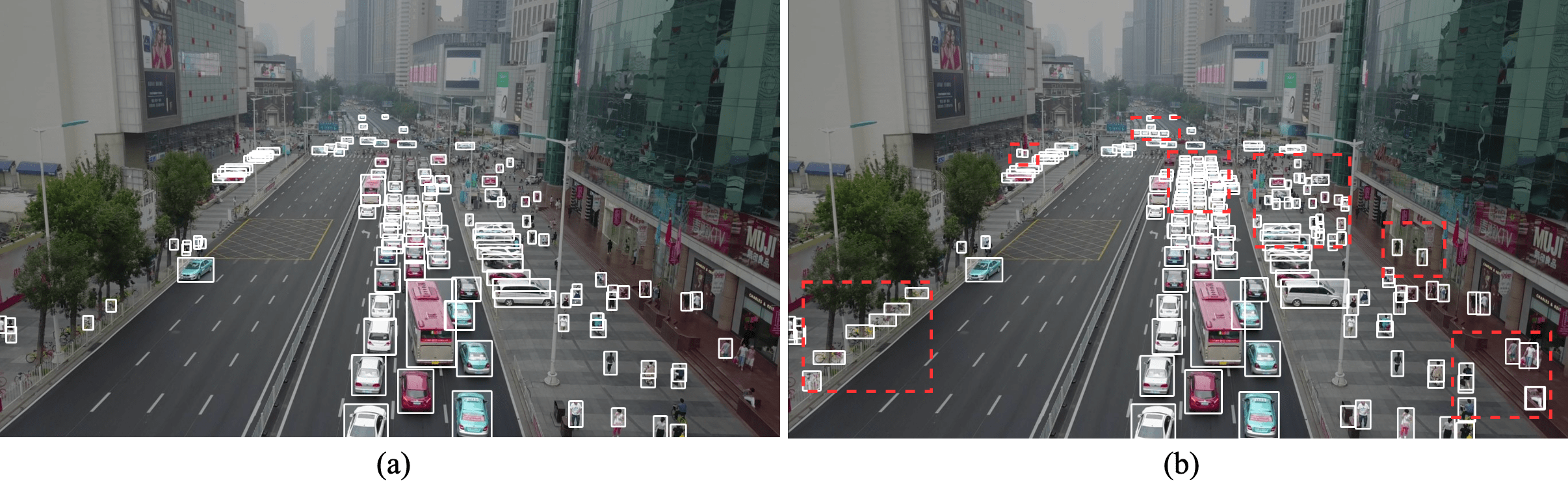}}
\caption{Inference results on a sample image with challenging targets on iOS deployment. (a) EDNet-Tiny, (b) EDNet-B. The labels and colors for bounding boxes are removed for better clarity to focus on the detection capabilities comparison. The red dotted boxes are manually added after predicted bounding boxes to highlight the superior performance of the bigger size model in regions with smaller and more crowded targets. We observe conspicuous stronger detection capabilities with the bigger model, notably smaller vehicles that are far.}
\label{fig:tb}
\end{figure*}

The results for inference speed, measured in milliseconds per image, are displayed in Fig.\ref{fig:mobile}. On the newest iPhone 15, models can run between 21 frames per second (FPS) and 69 FPS, providing robust performance. All seven models can also operate seamlessly on the iPhone 12, with EDNet-Tiny achieving 55 FPS and EDNet-X reaching 16 FPS. We achieve over 25 FPS with the EDNet-B model, ensuring real-time processing capabilities on average-performing mobile devices like the iPhone 12, where we also conducted the ablation study on this variant. Larger models are better suited for deployment on embedded GPUs, such as the NVIDIA Jetson, which are ideal for premium-grade UAVs. The detection performance of EDNet-Tiny and EDNet-B is compared on a challenging image containing small objects, as illustrated in Fig.\ref{fig:tb}. Here, we observe a conspicuously higher number of detected objects, such as bicycles on the left, pedestrians on the right, and notably more small cars in the distance, reflecting the trade-off between model size and accuracy, as previously presented in Fig.\ref{fig:benchmark} and Table\ref{tab:model_comparison}.

\section{Conclusion}

In this paper, we presented EDNet, a novel framework optimized for small target detection in UAV imagery. EDNet introduces key architectural innovations, including the uniquely designed C2f-FCA block for enhanced feature extraction with reduced computational complexity, and the XSmall detection head combined with the Cross Concat strategy for improved feature fusion, leading to gains in both accuracy and inference speed. The incorporation of WIoU also enhances bounding box regression. 

With seven variants ranging from tiny to XL, EDNet consistently outperforms all YOLO iterations, achieving up to a 5.6\% mAP@50 gain compared to YOLOv10. Notably, EDNet runs from 16 FPS to 55 FPS on an iPhone 12 in real time. We propose a tiny variant with superior efficiency, making it ideal for resource-constrained edge deployment scenarios. EDNet also surpasses SOTA YOLO models at larger sizes while using significantly fewer parameters. Overall, EDNet provides a powerful and efficient solution for UAV-based small target detection, balancing accuracy, computational efficiency, and adaptability across diverse hardware platforms. 

Although all sizes of EDNet's efficiency is proved on iPhone, limitations include for example the deployment of larger sizes of EDNet on resource-constrained embedded systems. Future work includes further making the model more lightweight while preserving the accuracy to target lower-grade hardware and extending its application to remote-sensing datasets.

\end{document}